\documentclass[11pt]{article}

\usepackage[final]{acl}

\usepackage{times}
\usepackage{latexsym}

\usepackage[T1]{fontenc}

\usepackage[utf8]{inputenc}

\usepackage{microtype}

\usepackage{inconsolata}

\usepackage{graphicx}

\usepackage{booktabs} 
\usepackage[caption=false]{subfig}
\usepackage{color}
\usepackage{microtype} 
\usepackage{amsmath,amsfonts}
\usepackage{mathtools} 
\usepackage{mathrsfs}  
\usepackage{bm}
\usepackage{multirow}
\usepackage{makecell}
\usepackage{stfloats}

\newcommand{\redhl}[1]{{\color{red}#1}}
\newcommand{\cyanhl}[1]{{\color{cyan}#1}}
\newcommand{\magentahl}[1]{{\color{magenta}#1}}
\usepackage{CJKutf8}
\usepackage{arydshln}
\usepackage{enumerate}

\title{\textsc{Interactor}: Agentic RL oriented Iterative Creation for Ad Description Generation in Sponsored Search}

\author{Penghui Wei, Jiayu Wu, Chao Ye, Zhi Guo, Shuanglong Li \and Lin Liu \\
         Search Ads, Baidu Inc. \\ \texttt{\{weipenghui,wujiayu01,yechao01,guozhi,lishuanglong,liulin03\}@baidu.com}}

\begin{document}
\maketitle
\begin{abstract}
This paper focuses on automatically generating informative ad descriptions in sponsored search. Unlike ad titles which are usually optimized to attract user click feedbacks, ad descriptions have a longer text span and possess the potential of incorporating world knowledge to address user search intents while presenting the fine-grained selling points of the ads. We propose \textsc{Interactor}, a multi-turn iterative creation framework optimized with agentic RL for ad description generation. The generation model acts as a policy that interacts with a customized environment consisting of multiple generative reward models. Given initial generations by the policy, the customized GenRMs evaluate qualities including knowledge capacity and landing page consistency, providing both binary signals and detailed feedbacks. The policy then iteratively refines the descriptions based on such feedbacks to ensure continuous improvement. Experiments show that it significantly outperforms state-of-the-art ad text generation approaches in generating knowledge-rich and faithful ad descriptions. Since late May 2026, it has been deployed online in a leading search ads system, where the framework serves over 140k advertisers, contributing to both ad revenue and user experience.
\end{abstract}

\section{Introduction}

Search advertising plays a crucial role to bridge user intents and business services, which brings substantial revenue to search engine platforms. Advertisers usually put a lot of effort into designing ad creatives (e.g., title, description and image) to clearly show the advantages of their services and attract potential audiences. 
With the rise of generative models and AI generated content (AIGC) techniques, extensive research proposed to automatically generate ad texts and images to improve click-through rate (CTR), achieving better delivery performance compared to the creatives provided by advertisers. Most work on ad text generation has focused on ad titles, while ad descriptions -- which also occupy a large visual space and allow for more informative contents -- remain underexplored. 
An ideal ad description should exploit the superiority of longer text span, integrating both 1) relevant and appropriate world knowledge that matches user's search intent, and 2) fine-grained selling points mined from landing pages to faithfully represent the advantages of services or products. 

\begin{figure}[t]
\centering
\centerline{\includegraphics[width=1\columnwidth]{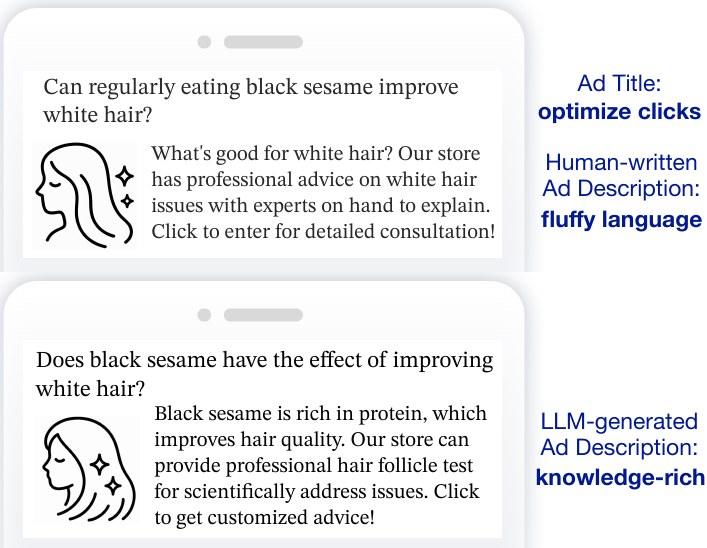}}
\caption{An example of ad description with query `Can black sesame turn white hair black?'. Compared to title, it has a longer span and should be more informative.}
\label{fig:example}
\end{figure}

Ad description generation is an open-ended task with no ground-truth reference to imitate. Reinforcement learning, which alternates between collecting rollouts for exploration and optimizing the policy with rewards for exploitation, is therefore a natural solution, especially in online advertising where user preferences change dynamically. 
The state-of-the-art approaches for ad text generation~\cite{chen2025ctr,wang2025beyond} learn to generate titles achieving better quality and higher CTR via preference optimization or RLHF, with implicit or explicit reward modeling to guide LLM training via scalar signals. 
However, the ad description generation task cannot directly adopt the above approaches heavily relying on scalar rewards during optimization. 
Compared to ad titles, ad descriptions are relatively long and must satisfy more quality constraints, thus the generated results inevitably suffer from various subtle inaccuracies. 
Current approaches struggle to produce detailed guidance information for revising incorrect generations, which results in inefficient exploration due to low-quality rollouts. 

To tackle the above challenges, in this paper we propose \textsc{Interactor}, an agentic RL oriented iterative creation framework for informative ad description generation.\footnote{\textsc{Interactor}: Agent\textbf{{i}}c Rei\textbf{{n}}forcemen\textbf{{t}} L\textbf{{e}}a\textbf{{r}}ning for \textbf{{A}}d Des\textbf{{c}}rip\textbf{{t}}i\textbf{{o}}n Gene\textbf{{r}}ation.} 
Compared to generating the final description in a single pass optimized by scalar rewards, 
the \textsc{Interactor} framework incorporates a multi-turn iteration process, in which the LLM to be optimized for description generation is regarded as an agent that receives detailed feedback from customized generative reward models to refine previous generations for improvement. 

Specifically, the LLM policy interacts with a customized environment consisting of GenRMs responsible for evaluating multi-dimensional qualities such as knowledge capacity and landing page consistency. 
After the initial descriptions generated by the policy, the GenRMs discriminate the generations with  binary signals and also provide detailed reasoning feedbacks (e.g., the reason why the initial generations miss appropriate world knowledge related to user intent, or pointing out the unfaithful contents inconsistent with landing pages), based on the predefined rubrics summarized from business requirements. 
If the criteria are not met, the policy then iteratively refines the descriptions via incorporating such feedbacks into context to form a multi-turn iterative creation process. The policy is optimized via agentic RL with multi-dimensional rewards to ensure that the quality of subsequent generations improves continuously. 

Since late May 2026, the \textsc{Interactor} framework has been deployed online in a leading search ads system, contributing to both ad revenue and user experience. The main contributions are:
\begin{itemize}
    \item We identify the necessity to shift from CTR-driven title generation to informative description generation in industrial practice, highlighting the importance of explicitly modeling world knowledge to satisfy user intent.
    \item We propose a multi-turn iteration framework \textsc{Interactor} for ad description generation, in which the LLM policy is optimized via agentic RL with customized generative rewards to achieve continuous improvement on knowledge capacity and landing page consistency.
    \item Experiments on industrial datasets show that the \textsc{Interactor} framework significantly outperforms state-of-the-art single-turn approaches in generating knowledge-rich and faithful ad descriptions. Online A/B test verifies its effectiveness in improving ad revenue and user experience.
\end{itemize}

\section{Problem Definition}
We introduce the task definition of ad description generation in sponsored search. Given a user search query $x_\mathrm{user}$ and an ad's landing page information $x_\mathrm{ad}$, the task's objective is to learn a generation policy $\pi_{\theta}(y \mid x_\mathrm{user},x_\mathrm{ad})$ that outputs an informative description $y$ satisfying quality requirements as well as achieving higher CTR, taking both user experience and advertiser value into account.\footnote{In practice, to provide more information as input, we also use the keyword (retrieved based on user query or recommended by the system) and ad metadata (e.g., advertiser name and trade name) to supplement $x_\mathrm{user}$ and $x_\mathrm{ad}$, respectively. } 

The quality is evaluated by two dimensions. 1) Knowledge capacity: as a long-form text (compared to ad title), the ad description $y$ is expected to contain world knowledge that is relevant to user search intent $x_\mathrm{user}$ for improving informativeness. 2) Landing page consistency: the ad description $y$'s detailed content that represents the advantages of services or products must be strictly entailed by the landing page information $x_\mathrm{ad}$, without any unfaithful exaggerations that mislead users.

\section{Methodology: The \textsc{Interactor} Framework}
Unlike current ad text generation approaches which generate the results in a single-turn process, 
the \textsc{Interactor} framework formulates the description generation process as \textit{multi-turn iteration} optimized with agentic RL, where the LLM policy interacts with a customized environment that provides detailed feedbacks about content quality for continuous improvement. 
Our framework can refine incorrect expressions with fine-grained feedbacks, which is suitable for open-ended, long-form ad description generation.

\begin{figure}[t]
\includegraphics[width=1\linewidth]{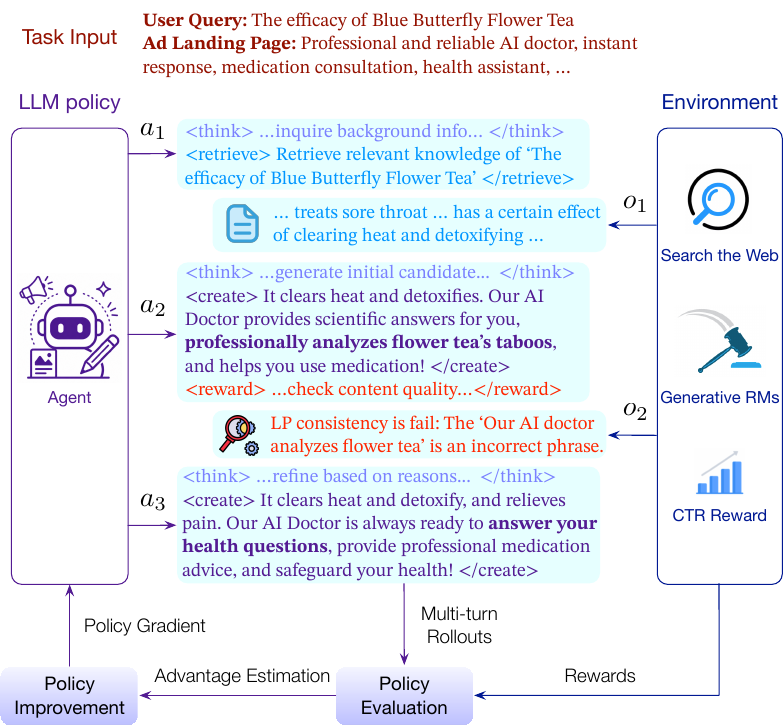} 
\caption {The overall workflow of the \textsc{Interactor} framework for generating informative ad descriptions.}
\label{fig:multi-turn}
\end{figure}

\subsection{Formulation: Multi-Turn Iteration}
Figure~\ref{fig:multi-turn} shows the overview of \textsc{Interactor}, where an LLM policy interacts with the customized environment to produce the final generation.

Formally, the initial context consists of customized task prompt $x_\mathrm{task}$ and inputs $\{x_\mathrm{user}, x_\mathrm{ad}\}$. 
The task prompt $x_\mathrm{task}$ strictly structures the LLM policy’s output format of each turn into two parts: 
\begin{enumerate}
    \item A thinking process (enclosed in special tokens \texttt{<think>} \texttt{</think>}) that understands context including task inputs and environment's observations from previous turns;
    \item A response that reflects a real \textbf{action}, which may \textit{request the search engine} (enclosed in \texttt{<retrieve>} \texttt{</retrieve>}) for injecting relevant world knowledge, \textit{generate an ad description} based on all turns' contexts (enclosed in \texttt{<create>} \texttt{</create>}), and \textit{request reward models} (enclosed in \texttt{<reward>} \texttt{</reward>}) for fine-grained evaluation and refinement.
\end{enumerate}

At the $t$-th turn, the LLM policy takes an \textbf{action} $a_t$ (thinking and response) based on the context $c_t =([x_\mathrm{task}; x_\mathrm{user}; x_\mathrm{ad}], a_1, o_1, \ldots, a_{t-1},o_{t-1})$,\footnote{For the first turn, the context $c_1 = ([x_\mathrm{task}; x_\mathrm{user}; x_\mathrm{ad}])$.} then receives an \textbf{observation} $o_t$ (e.g., retrieved knowledge and reasoning information of rewards) from the environment and also a \textbf{reward} $r_t$ (multi-dimensional evaluations of qualities and CTR) if the action is generation. The objective is to optimize the policy $\pi_{\theta}(y \mid x_\mathrm{user},x_\mathrm{ad})$ guided by the reward signals with policy gradient.

\subsection{The Customized Environment for Ad Description Generation}

To perform multi-turn iteration, we design a customized environment that provides observations to the learned policy through generative reward models and a knowledge retrieval engine.

\subsubsection{Rewards: Hybrid of Rubric-based Generative RMs and Human Feedback}
\label{sec:rewards}
\textbf{Rubric-based GenRMs for Quality.} To evaluate the qualities and also guide further refinement, we build two generative reward models for knowledge capacity and landing page consistency -- the main dimensions of content quality -- based on predefined rubrics to transform open-ended evaluation to verifiable signals~\cite{kim2024prometheus,gunjal2025rubrics}. Unlike current state-of-the-art ad text generation approaches which only guide RLHF via scalar rewards, our GenRMs produce both binary signals and fine-grained reasoning information. 

Beyond scalar rewards for exploitation, the detailed feedbacks facilitate next turn's refinement of ad descriptions, directly contributing to high-quality rollouts for efficient exploration. 
As shown in Figure~\ref{fig:multi-turn}, for landing page consistency, the GenRM points out unfaithful phrases that mislead users. For knowledge capacity, it can return the reason why the initial description misses appropriate world knowledge that matches user intent. 

Specifically, for each quality dimension, we first summarize the rubrics consisting of multiple verifiable rules according to business requirements, and then prompt LLMs with few-shot in-context learning to obtain 1) binary result of the reward, if and only if all rubrics are satisfied then the quality reward is 1, and 2) fine-grained reasoning information that explains the improvement directions for a given description. 
The predefined rubrics, prompts and other details of the GenRMs can be found in Appendix~\ref{sec-appendix:details}.

\noindent\textbf{CTR Estimation with Human Feedback.} To evaluate the attractiveness of generated descriptions, we build a CTR estimation model trained on the online logs of user feedbacks in our search ads system. It serves as a reward that produces a real number $r_t^{\mathrm{CTR}}\in(0, 1)$ as the predicted CTR given both task inputs and a generated description. 

To capture the dynamics of user preference, the CTR estimation model is updated incrementally in a daily fashion. For each day we continually train the model via loading previous day's checkpoint and optimizing parameters with fresh data.

\subsubsection{Observations: Reasoning from GenRMs and Knowledge from Search Engine}

The description generation model acts as an LLM policy $\pi_{\theta}(\cdot\mid c_t)$ that interacts with the environment to receive observations to augment contexts for further actions. 
The text observation $o_t$ is typically the reasoning information of GenRMs for quality evaluation and refinement, or the retrieved knowledge from the search engine for generating contents that satisfy user search intent. 

\noindent\textbf{Reasoning from Quality GenRMs.} For a turn with the context $c_t$, if the policy's action $a_t$ is to generate a description, the GenRMs will be requested to provide both scalar signals and reasoning information with fine-grained feedbacks about incorrect phrases in the description. 
Such reasoning information is used as observation $o_t$ to augment next turn's context $c_{t+1}$, guiding the policy to inject missing world knowledge or modify unfaithful contents that are inconsistent with the landing page.

\noindent\textbf{Retrieved Knowledge from Search Engine.} 
The policy's action $a_t$ may request the search engine (triggered by generating content in \texttt{<retrieve>} \texttt{</retrieve>}) to retrieve relevant knowledge based on the search query $x_{\mathrm{user}}$ (usually at the first turn), which helps the generated ad description contain appropriate world knowledge that matches user search intent. Such retrieved knowledge is regarded as observation $o_t$ to augment next turn's context $c_{t+1}$, which supplements the landing page information to enhance the task input.\footnote{Although the base LLM of the generation model stores world knowledge implicitly in the parameters, in practice we found that requesting our in-house search engine to explicitly enhance the context can achieve better generation results. }  

Specifically, we use our in-house search engine, which indexes a massive corpus of encyclopedic web pages and thus is highly well-suited for acquiring world knowledge. Given the search query $x_{\mathrm{user}}$, the document of the top-1 retrieved web page is used as the knowledge to augment context.

\subsection{Agentic RL for Optimizing Description Generation Policy}

For each turn $t$, the description generation model takes an action $a_t\sim\pi_{\theta}(\cdot\mid c_t)$ based on the context $c_t$ containing task inputs, previous actions and observations. 
It is expected that at subsequent turns the policy can generate descriptions with better quality and higher CTR given the fine-grained reasoning from the environment. 

\subsubsection{Multi-Turn Rollouts}
Based on the iterative creation nature, each episode constitutes a multi-turn rollout sequence $\tau=([x_\mathrm{task}; x_\mathrm{user}; x_\mathrm{ad}], a_1, o_1, r_1, \ldots, a_{T}, o_T, r_T)$, where each action's token sequence structure contains a thinking sequence and a response sequence~\cite{yao2023react}, and the \textit{final generated description} of the LLM policy is extracted from the response part of the last turn's action $a_T$. 

Specifically, at subsequent turns, the LLM policy generates descriptions that refine incorrect phrases with the help of the reasoning information from the environment. As shown in Figure~\ref{fig:multi-turn}, the subsequent turn successfully refines the initial description via modifying unfaithful phrase and also injecting knowledge guided by the detailed feedback, achieving improvement via iterative creation. 

Here the reward signal $r_t$ is the weighted sum of quality rewards and predicted CTR:
\begin{equation}
    r_t = \sum_{*\in\{\mathrm{KC}, \mathrm{LP}, \mathrm{CTR}\}} w^*\cdot r_t^{*}
\end{equation}
where $r_t^{\mathrm{KC}}$ / $r_t^{\mathrm{LP}}$ denote the binary signal from the GenRM of knowledge capacity / landing page consistency, and $r_t^{\mathrm{CTR}}$ denotes the predicted CTR, with their factors $w^*$.

\subsubsection{Optimization}
We use group sequence policy optimization (GSPO)~\cite{zheng2025group} to learn the LLM policy, which is critic-free~\cite{shao2024deepseekmath} and defines a sequence-level importance ratio to match the rewarding granularity. Formally, we maximize:
\begin{equation}
    \mathop{\mathbb E}\limits_{\{\tau^{(i)}\}_{i=1}^G\sim\pi_{\theta_{\mathrm{old}}}}\left[  \frac1G \sum_{i=1}^G \left(\frac{\pi_{\theta}(\tau^{(i)})} {\pi_{\theta_{\mathrm{old}}}(\tau^{(i)})}\right)^{\frac{1}{|\tau^{(i)}|}} A^{(i)}_{\mathrm{norm}}   \right]
\end{equation}
Here we omit detailed clip operation and KL term for brevity. The advantage $A^{(i)}_{\mathrm{norm}}$ of each rollout sequence $\tau^{(i)}$ per prompt is computed using the last turn's reward $r_T^{(i)}$ via group-based estimation of $G$ rollouts and global normalization in each training step's data $\mathcal D_{\mathrm{step}}$~\cite{hu2025reinforce}:
\begin{equation}
    A^{(i)}_{\mathrm{norm}} = \mathsf{Norm}\Big(\mathsf{Norm}(r_T^{(i)}\mid i\in [1,G]) \mid \mathcal D_{\mathrm{step}}\Big)\,.
\end{equation}
Note that in a rollout sequence $\tau$, a token may be generated by the LLM policy (i.e., $a_t$), provided by the environment (i.e., $o_t$), or come from the task input (i.e., $[x_\mathrm{task}; x_\mathrm{user}; x_\mathrm{ad}]$). Thus during policy gradient computation, we only reserve action tokens while masking the others.

\section{Experiments}

\subsection{Experimental Setup}
\noindent\textbf{Dataset Construction}\quad 
To our knowledge, no publicly available benchmark exists for ad description generation,\footnote{Current open-source benchmark of ad text generation focuses on ad title~\cite{mita2024striking}.} thus we verify the effectiveness of \textsc{Interactor} on a large-scale proprietary dataset collected from our search ads system, operating as a core component of a leading search engine in Chinese. We sample the impression log of search ads to obtain 1 million user-ad pairs as the training set, and hold-out 5k / 10k pairs as the validation / test set where the impression time is strictly later than training data. 

\paragraph{\textbf{Evaluation Metrics}}
We thoroughly evaluate the generations with five dimensions: 1) {\underline{Basic quality}}: measures whether the content contains basic errors such as wrongly-written characters / punctuations and repetitive phrases. 2) {\underline{Informativeness}}: indicates whether the description contains knowledge relevant to user query. 3) {\underline{Faithfulness}}: indicates the landing page consistency. 4) {\underline{Attractiveness}}: reflects whether the user intends to click on. 5) {\underline{Diversity}}: measures the content similarity across multiple candidates. 
More details of the evaluation process are listed in Appendix~\ref{sec-appendix:evaluation}.

\paragraph{\textbf{Competitors}}
We use the state-of-the-art approaches of ad text generation, as well as our best-performing production model as baselines:
\begin{itemize}
    \item \textsc{Ctop}~\cite{chen2025ctr} and \textsc{Diver}~\cite{wang2025beyond}: the SOTA approaches for ad title generation via preference optimization and RLHF respectively to jointly optimize quality and CTR. We directly adopt them to perform description generation task.
    \item \textsc{Single-turn RL}: the best-performing production policy of our system that incorporates the same knowledge retrieval and reward design as \textsc{Interactor}. It is optimized via single-turn RL and thus generates the final description without iteration.
\end{itemize}

\noindent\textbf{Implementation Details}\quad For fair comparison, we unify the base model as \texttt{Qwen3-30B-A3B}~\cite{yang2025qwen3}, and employ GSPO for all RL-based competitors. We choose \texttt{Slime}\footnote{\url{https://github.com/THUDM/slime}} and \texttt{SGLang}~\cite{zheng2024sglang} as the post-training and inference framework, respectively. Appendix~\ref{sec-appendix:implementation} lists more details.

\begin{table}[t]
\renewcommand{\arraystretch}{1.3}
\centering
\footnotesize  
\begin{tabular}{p{6.5em}p{2em}p{2em}p{2em}p{2em}p{2em}}
\toprule
\textbf{Approach} & Basic. & Info. & Faith. & Attract. & Div. \\ 
\midrule
\textsc{Ctop}        & 0.989 & 0.274          & 0.634          & 7.8\%           & 0.662 \\
\textsc{Diver}       & 0.985 & 0.359          & 0.724          & 11.0\%          & 0.684 \\
\cdashline{1-6}[2pt/2pt]
\textsc{Single-turn} & 0.986 & 0.693          & 0.726          & 12.3\%          & 0.715 \\
\textsc{Interactor}  & 0.986 & \textbf{0.715} & \textbf{0.872} & \textbf{12.4\%} & \textbf{0.732} \\
\bottomrule
\end{tabular}  
\caption{Main results of basic quality, informativeness, faithfulness, attractiveness and diversity.}
\label{results:main}
\end{table}

\subsection{Main Results}

Table~\ref{results:main} reports the evaluation results of all compared approaches. For each input we generate ten candidates, and report $\mathrm{Average@10}$ as the results. 

The superiority of RL-based approaches compared to \textsc{Ctop} shows that learning from pre-collected static data with preference optimization is sub-optimal in open-ended ad description generation task, especially for the attractiveness metric which is measured by predicted CTR that reflects user's dynamic interests. For the informativeness metric, both \textsc{Interactor} and our production model \textsc{Single-turn RL} outperform other competitors by a large margin, demonstrating that explicitly incorporating world knowledge is crucial for generating high-quality descriptions that satisfy user intent, which reveals the essential difference between ad description and ad title. 

The \textsc{Interactor} framework significantly outperforms other competitors on the whole, especially for the faithfulness metric which is crucial but difficult to optimize for ad text. The results verify the effectiveness of the multi-turn iterative creation for optimizing landing page consistency with the help of fine-grained reasoning information, and also indicate that the optimization paradigm of agentic RL possesses the ability of improving various metrics simultaneously.

\subsection{Discussion}

\paragraph{\textbf{Effects of Multi-turn Iteration}}
During the inference stage of \textsc{Interactor}, we also extract the generated description from the penultimate turn  and report the metrics in Table~\ref{results:ablation}. We can see that the last turn beats the penultimate turn on most metrics, demonstrating that the iterative creation process brings continuous improvement on description quality. Moreover, the penultimate turn still outperforms \textsc{Single-turn RL}, which shows that the agentic RL framework indeed learns to generate better descriptions rather than merely suppressing previous turns to inflate the final improvement.

\paragraph{\textbf{Ablation of Reasoning for Rollouts}}
We further verify the effectiveness of the detailed reasoning information from the environment. Specifically, we train an LLM policy which omits the guide of reasoning information from the GenRMs and performs multi-turn rollouts based on the binary signals only. As the results of ``w/o reasoning'' in Table~\ref{results:ablation}, the performance drops dramatically, which reveals that the decisive factor of \textsc{Interactor}'s iterative creation process is the fine-grained feedbacks of the GenRMs and indicates the effectiveness of the overall multi-turn framework.

\paragraph{\textbf{Sensitivity to Model Choice of GenRMs}}
Based on the above results, we further analyze how sensitive the performance is to the choice of GenRM backbone. Specifically, we replace the GenRMs from the \texttt{Qwen3-30B-A3B} to the smaller-sized \texttt{Qwen3-8B}~\cite{yang2025qwen3} with the same prompts. The results of ``smaller GenRMs'' in Table~\ref{results:ablation} verify that our framework is not very sensitive to the choice of GenRMs, however stronger GenRMs can further improve the performance of generated descriptions on multiple metrics.

\begin{table}[t]
\renewcommand{\arraystretch}{1.3}
\centering
\footnotesize 
\begin{tabular}{lp{2em}p{2em}p{2em}p{2em}p{2em}}
\toprule
\textbf{Approach} & Basic. & Info. & Faith. & Attract.   \\  
\midrule
\textsc{Interactor}  & \textbf{0.986} & \textbf{0.715} & \textbf{0.872} & 12.4\%  \\
\cdashline{1-5}[2pt/2pt]
\quad penultimate    & 0.982          & 0.694          & 0.798          & \textbf{12.5\%} \\
\quad w/o reasoning  & 0.985          & 0.636          & 0.706          & 12.3\%    \\
\quad smaller GenRMs & 0.985          & 0.692          & 0.861          & 12.3\%    \\ 
\bottomrule
\end{tabular}  
\caption{Ablation studies of multi-turn iteration, reasoning, and GenRM choice for \textsc{Interactor}.}
\label{results:ablation}
\end{table}

\subsection{Case Analysis}
In Table~\ref{results:case} we qualitatively show a generated case. We observe that the subsequent turn successfully refines the initial description via modifying unfaithful phrase guided by the detailed feedback from the customized GenRMs and injecting knowledge relevant to user query, achieving a significant quality improvement via reasoning-guided iterative creation. This also illustrates that the phenomenon where the penultimate turn achieves a slightly higher CTR than the final turn is not a flaw, but the intended behavior of our policy: in earlier turn the model may generate attractive but unsupported claims, then the later turn utilizes fine-grained feedbacks to explicitly suppress these unfaithful elements, replacing them with factual ones.

\begin{table}[t]
\centering
\scriptsize
\begin{tabular}{p{6em}p{23em}}
\toprule
\multirow{2}*{\textbf{Input}} & {\textbf{User Query}: \begin{CJK*}{UTF8}{gbsn}电钢琴型号推荐\end{CJK*} (Recommended electric piano models)}\\
& \textbf{Landing Page}: \begin{CJK*}{UTF8}{gbsn}家用电钢琴各种型号,雅马哈/美得理厂家直销原装联保,便携式可折叠/重锤,考级/初学/儿童/成人,送上门...\end{CJK*} (Various home electric pianos, Yamaha/Medeli factory direct sales, original packaging and warranty, portable and foldable/heavy hammer, examination/beginner/children/adult, door-to-door delivery...) \\  
\midrule 
Action $a_1$  & \begin{CJK*}{UTF8}{gbsn} 检索“电钢琴型号推荐”相关知识 \end{CJK*} (Retrieve relevant knowledge of 'Recommended electric piano models') \\
Observation $o_1$   & \begin{CJK*}{UTF8}{gbsn}...初学者买电钢,首先{\magentahl{考虑口碑和声誉}},选购时需亲自{\magentahl{体验手感}},音质也是不能忽略的因素...\end{CJK*} (...For beginners of purchasing an electric piano, it is important to first {\magentahl{consider its reputation and word-of-mouth}}, essential to personally {\magentahl{experience the feel}} of the keys. The sound quality is also a factor that cannot be ignored...)\\
\cmidrule(lr){1-2}
Action $a_2$ & \begin{CJK*}{UTF8}{gbsn}电钢琴型号推荐!雅马哈等品牌热销,重锤键盘专业级手感,初学考级皆适用.原装正品,{\redhl{{快递包邮}}},便携折叠款可选.限时优惠购!\end{CJK*} (Recommended electric piano models! Popular brands such as Yamaha, featuring heavy-hammer keyboards for professional-level touch, suitable for beginners and grade exams. Genuine products, {\redhl{{free shipping}}} via express delivery, and portable folding models available. Limited-time discount!)\\
Observation $o_2$ & \begin{CJK*}{UTF8}{gbsn}知识化程度不通过：检索到电钢选购知识,但未体现在描述中｜落地页一致性不通过：卖点"包邮"的原意及类似含义未出现在广告信息中。\end{CJK*} (Knowledge capacity is fail: Knowledge about electric piano was retrieved, but it was not integrated. | LP consistency is fail: The selling point of `free shipping' mentioned in the candidate is not found in the ad information with a similar meaning.)\\
\cmidrule(lr){1-2}
Action $a_3$  & \begin{CJK*}{UTF8}{gbsn}新手选购电钢需{\magentahl{兼顾键盘质量,手感及品牌口碑}}!热销雅马哈等品牌,重锤键盘模拟真实手感,适合初学考级.原装联保,{\cyanhl{快递上楼}},便携折叠设计,限时优惠!\end{CJK*} (For beginners of electric pianos, it's essential to {\magentahl{consider keyboard quality, feel, and brand reputation}}! Popular brands like Yamaha offer heavy-hammer keyboards simulating real touch, ideal for beginners and grade exams. With original warranty, {\cyanhl{door-to-door delivery}}, portable folding design, and limited-time discounts!)\\
\bottomrule
\end{tabular}
\caption{Case analysis. We omit the sequence enclosed in \texttt{<think>} \texttt{</think>} for brevity. Contents in the parentheses are the results translated to English.}
\label{results:case}
\end{table}

\subsection{Online A/B Test and Human Evaluation}
We conduct online A/B test in our search ads system, and the experiment group covers over \textbf{140k advertisers} and accumulated over \textbf{60 million ad impressions}. Results verify that \textsc{Interactor} contributes to both ad revenue (\textbf{+0.74\%}) and user experience (top-grade rate \textbf{+7pp} of content quality in human evaluation). 
Since late May 2026, it has been deployed online to serve the main traffic.

\paragraph{\textbf{Deployment}}
The descriptions from \textsc{Interactor} are generated with offline batching inference and cached in a KV database for rapid retrieval during real-time auction. 
Overall, the dynamic creative module in our system consists of a generation stage and a ranking stage. After the ad pre-ranking module, the creative generation stage assembles candidates via 1) retrieving advertiser-uploaded assets and offline-generated titles, descriptions and images from KV databases; and 2) generating titles in real time~\cite{dong2026efficient}, where this real-time branch is selectively deployed only for high-value traffic. 
Subsequently, the creative ranking stage performs personalized scoring at the creative combination level, selecting the top-1 combination to downstream ad ranking module.\footnote{All offline generated creatives are fed into a risk-control process to delete unsecured results before being stored to KV database, and real-time generated titles are fed into an online risk-control stage before the ranking stage.} Because multiple offline branches exist for each creative type, a single ad can yield thousands of potential combinations.  
During experiment, the candidate size allocated to the \textsc{Interactor} branch is two per ad, and concurrently we deprecate an old branch.

\paragraph{\textbf{Online A/B Test}}
For online A/B test, the randomization unit is at the search page view (PV) level. 
Our experiment achieved \textbf{+0.74\%} lift on ad revenue (where the 95\% confidence interval is [0.13\%,1.40\%] with p-value 0.02). Guardrail metrics, such as the App's daily active users (DAU), search PV per capita and lifetime value (LTV), as well as the system's p99 latency and resource overhead for infrastructure, showed no significant change in our experiment.

\paragraph{\textbf{Human Evaluation}}
Human evaluation for user experience is a crucial step before deploying a new algorithm online. 
We have a strict in-house criterion that defines top-grade description. For experiment and control groups, we randomly sample 400 descriptions per group for evaluation. 
Results show that the generated descriptions of \textsc{Interactor} achieve \textbf{+7pp} on top-grade rate, demonstrating that our policy achieves better user experience compared to previous production policy.

Beyond this, our ads system undergoes continuous, large-scale monthly human evaluation, where randomly-sampled impression creatives are collected to perform human evaluation for monitoring long-term quality. The \textsc{Interactor} was deployed to serve the main traffic in the last week of May 2026, and the routine evaluation in June 2026 demonstrated a sustained quality improvement of +9pp on top-grade rate compared to the first three weeks in May 2026. This robustly proves that \textsc{Interactor} improves true creative quality.

\section{Conclusion}
We identify the necessity to shift from generating CTR-driven ad titles to informative ad descriptions in industrial practice, highlighting the importance of explicitly modeling world knowledge to satisfy user intent. We propose \textsc{Interactor}, a multi-turn iteration framework optimized via agentic RL to achieve continuous improvement on multi-dimensional evaluations. Online A/B test verifies its effectiveness in improving ad revenue and user experience for an industry-scale system.

\section*{Ethical Considerations}
When we apply AIGC technologies for online advertising business, both content security and user privacy protection are crucial. 
We make the following efforts to ensure only secure contents will be pushed online and alleviate potential privacy risks: 1) We only generate ad creatives for the advertisers who have submitted authorization to our advertising platform. 2) We have a strict risk-control process before deployment, which ensures the content security via a large number (over 100) of rules and high performing models to delete harmful results and rule-breaking contents, covering the recognition capacity for various trivial details about negative content. 3) Our algorithm does not incorporate any user privacy information (such as age and gender), which is not collected in the dataset. 

To substantiate the safety of deploying \textsc{Interactor}, we provide additional details about the risk-control process and its empirical performance, especially to sensitive categories.

We first report the faithfulness and risk-control deletion rates for sensitive categories. 
On the hold-out test set, the $\mathrm{Average@10}$ faithfulness scores of \textsc{Interactor} are 0.859 for the health category and 0.923 for the financial service category. During offline generation, the daily deletion rates by the risk-control process are 4.3\% for the health category and 2.7\% for the financial service category. These results characterize the quality of generated outputs and the behavior of the downstream risk-control process on sensitive categories.

We further report the effectiveness of the risk-control process for \textsc{Interactor} outputs. Before the online experiment, we accumulated over 10k samples that were labeled as containing phrases unsuitable for display (e.g., including exaggerated statements inconsistent with the landing page) by human evaluation. We passed these samples through the risk-control process, and none of the samples remained, corresponding to an observed residual rate of 0\%. This provides direct evidence that the risk-control process effectively identifies unsafe outputs represented in \textsc{Interactor}.

Regarding category-level restrictions, the risk-control algorithm applies stricter criteria to sensitive categories. For example, it prohibits potentially unsupported claims about treatment, diagnosis, or medication without sufficient evidence in health ads, as well as unsupported claims concerning investment, lending, or credit outcomes in financial ads. The risk-control rules and models are continuously updated based on new patterns of negative content identified through routine human evaluation and reports from advertisers and users. The same risk-control process covers both LLM-generated and advertiser-uploaded creatives.

We also maintain a real-time intervention pathway that can directly process reports from advertisers and users and remove unsafe content. On the advertiser side, for example, advertisers may submit unreasonable keywords to compete with competitors, which can trigger intervention. On the user side, an appeal button is provided at the ad display position for users to report problematic ads. Our system can process these reports and perform real-time intervention when necessary.

Therefore, the above safeguards in our system build a comprehensive protection mechanism: 1) strict criteria to all categories especially for sensitive ones, 2) dedicated risk-control rules and models applied to both generated and advertiser-uploaded creatives, 3) effective risk-control process covering each search request, and 4) post-deployment monitoring and real-time intervention.

\section*{Limitations}
Our proposed framework focuses on automated ad description generation that provides high-quality candidates to advertisers, however the corpora of advertiser-written descriptions are not explicitly exploited during our framework development (they only implicitly contribute to the development of GenRMs). 
We believe that our environment possesses the ability of providing editing suggestion for them, and thus how to mine the value of human-written corpora and build a human-in-the-loop optimization paradigm for ad text generation is a promising direction for further research.

\section*{Acknowledgments}
We thank all the anonymous reviewers for their valuable comments. We also thank Yi Zhang for his kind assistance.

\bibliography{custom}

\appendix

\section{Additional Results on Methods with Off-the-shelf LLMs}
During early experiments, we also evaluate frontier LLMs under few-shot and test-time scaling settings to explore the potential of multi-turn iteration. 
\begin{itemize}
    \item \textsc{Few-shot}~\cite{fewshot2020}: We prompt the \texttt{Gemini-3.1-Pro}\footnote{\url{https://deepmind.google/models/model-cards/gemini-3-1-pro/}} with detailed rubrics  of multi-dimensional quality requirements and few-shot examples to generate descriptions.
    \item \textsc{ReAct}~\cite{yao2023react}: We prompt the Gemini-3.1-Pro to iteratively critique and improve its own generations, without the explicit reward signals used in \textsc{Interactor}.
    \item \textsc{ReAct} w/ Customized Env.: It further incorporates the search engine and GenRMs used in the environment of \textsc{Interactor} to provide knowledge and feedback during iteration.
\end{itemize}
We also compare the test-time scaling settings of the SFT checkpoint:
\begin{itemize}
    \item \textsc{SFT} w/ Best-of-$N$: It first generates $N=20$ candidates with the SFT checkpoint of \textsc{Single-turn RL}, and then uses our reward signal to select 10 best-performing outputs. 
\end{itemize}

\begin{table}[t]
\renewcommand{\arraystretch}{1.3}
\centering
\footnotesize  
\begin{tabular}{p{10em}p{2em}p{2em}p{2em}p{2.2em}}
\toprule
\textbf{Approach} & Basic. & Info. & Faith. & Attract.  \\ 
\midrule
\textsc{Few-shot}    & 0.985 & 0.521 & 0.564 & 7.1\% \\
\textsc{ReAct}       & 0.985 & 0.526 & 0.589 & 6.9\% \\
\textsc{ReAct} w/ Custom. Env.  & 0.984 & 0.631 & 0.683 & 6.7\% \\
\cdashline{1-5}[2pt/2pt]
\textsc{SFT} w/ Best-of-$N$  & 0.986 & 0.685 & 0.711 & 11.6\% \\
\bottomrule
\end{tabular}  
\caption{Additional results on off-the-shelf LLMs under few-shot and test-time scaling settings, as well as the SFT checkpoint.}
\label{results:additional}
\end{table}

Table~\ref{results:additional} verifies that \textsc{Interactor} outperforms these baselines by a large margin, especially on attractiveness which reflects user dynamic preference.
User behaviors in search ads are highly dynamic and time-varying, and an LLM relying on static few-shot prompts cannot capture such dynamics and lacks alignment. The improvement on informativeness and faithfulness through incorporating customized environment further demonstrates the necessity of constructing accurate reward signals based on quality requirements.

\section{Implementation Details}
\subsection{Hybrid Rewards in the Environment}
\label{sec-appendix:details}
We build the GenRMs of knowledge capacity and landing page consistency to transform open-ended evaluation to verifiable signals. 
We first summarize the rubrics and then prompt \texttt{Qwen3-30B-A3B}~\cite{yang2025qwen3} to obtain binary result of the reward and fine-grained feedback as reasoning information. Table~\ref{details:prompts} lists the predefined rubrics and prompts of the GenRMs (without few-shot examples for brevity). The prompt is optimized via a continual check of the consistency between the GenRM and human annotators, and finally the consistency rate is above 0.8 on 200 randomly sampled examples, verifying their usability.

For the CTR estimation model, we use the \texttt{Qwen3-Embedding-0.6B}~\cite{zhang2025qwen3} as base model with an extra classification head to discriminate click or unclick,\footnote{We do not employ the token probability of `yes' or `no' because a pointwise CTR prediction model should meet the calibration condition, which produces probabilistic predictions reflecting true likelihoods.} where the input contains both $x_{\mathrm{user}},x_{\mathrm{ad}}$ and a generated description. We then perform continual training on the user click logs of our search ads system. The AUC on test set is above 0.7 and the PCOC (predicted CTR over the true CTR) is 1.04, verifying its usability as an offline model.

\subsection{Evaluation Metrics}
\label{sec-appendix:evaluation}
To evaluate basic quality, informativeness and faithfulness of generated descriptions, we employ the LLM-as-a-Judge method~\cite{zheng2023judging} via prompting \texttt{DeepSeek-V3.2}~\cite{liu2025deepseek} to return binary results. For basic quality, Table~\ref{details:prompts_basic} shows the prompt. For informativeness and faithfulness, we use the same prompts of the GenRMs for knowledge capacity and landing page consistency to perform evaluation, as shown in Table~\ref{details:prompts}.  

To evaluate attractiveness, we train a larger model based on \texttt{Qwen3-Embedding-8B}, using the same procedure as the CTR estimation model in policy learning.

To evaluate the diversity of multiple generations, we use $1 - \mathrm{PairwiseBLEU}$~\cite{shen2019mixture} that measures the dissimilarity across $K$ candidates $\{y_i\}_{i=1:K}$ per input ($K=10$ in our experiment):
\begin{equation}
    \mathrm{PairwiseBLEU} = \frac{1}{K(K-1)} \mathop\sum\limits_{i \neq j} \mathrm{BLEU}(y_i, y_j)\,.
\end{equation}

\subsection{Policy Learning and Inference}
\label{sec-appendix:implementation}
We use \texttt{Slime} for training with 8 $\times$ NVIDIA H800 GPUs, and the base model is \texttt{Qwen3-30B-A3B}. 
For the environment, we deploy a remote server for the reward models, and request our in-house search engine with internal API. The maximum turn is set to 3, and the group size is 8. The factors of multiple rewards are set to 1, 2 and 5.
We configure fully asynchronous to acceleration and employ routing replay~\cite{zheng2025group} to stabilize MoE training. 
For fair comparison, all competitors use the same set of 100k synthesized descriptions by \texttt{DeepSeek-V3.2} for cold-start SFT. Table~\ref{details:task} lists the task prompt of \textsc{Interactor}. 

We use \texttt{SGLang}~\cite{zheng2024sglang} for efficient offline batching inference. For each input we generate multiple candidates with top-$p$ sampling~\cite{holtzman2020topp}.
Although it seems that \textsc{Interactor} increases inference cost compared to the baseline, its computational resource utilization has been improved in industrial deployment. In the baseline deployment, descriptions are first generated and then filtered by post-processing models, therefore the compute spent on the discarded results is wasted. The \textsc{Interactor} essentially transfers post-processing into multi-turn iteration and eliminates extra post-processing, ultimately achieving more qualified results.

\begin{table*}[htbp]
\centering
\scriptsize
\begin{tabular}{p{9em}p{53em}}
\toprule
\multirow{1}*{\textbf{Quality Dimension}} & {\textbf{Prompt}} \\  
\midrule 
\textbf{Knowledge Capacity}   &  
    {\#\#\# Evaluation Criteria} \newline
    Given the ad landing page information, the retrieved content via searching web pages based on the user query, and the candidate description, please evaluate strictly based on the following rules and provide a judgment of "pass" or "fail". \newline
    \newline
    {\#\#\# Dimension: Knowledge Capacity} \newline
    1. The definition of world knowledge is: any knowledge that meets any of the following conditions: \newline
    \hspace*{1.5em}\textbullet\ Explain professional knowledge such as definitions, backgrounds, principles, and concepts (e.g., introduce "what is xx", "why is xx", "what are the symptoms of xx disease", etc.). \newline
    \hspace*{1.5em}\textbullet\ Describe the characteristics / attributes / applicable scenarios / mechanism of action / working principle / efficacy, etc. of things. \newline
    \hspace*{1.5em}\textbullet\ Popularize the objective impact of objects/behaviors. \newline
    \hspace*{1.5em}\textbullet\ Provide a neutral reference (e.g., "Usually, solutions to such problems include categories A, B, and C"). \newline
    \newline
    {\#\#\# Execution steps} \newline
    {Step 1}: Confirm whether the candidate description contains **knowledge from the retrieved content** \newline
    {Step 2}: If there is **{knowledge from the retrieved content}**, determine: \newline
    \hspace*{1.5em}Rule 2.1: The contained knowledge meets the definition of world knowledge \newline
    \hspace*{1.5em}Rule 2.2: The knowledge is consistent with the services that can be provided on the landing page \newline
    \hspace*{1.5em}The knowledge capacity is considered to be passed **only when both of the above rules are met** \newline
    {Step 3}: If there is **{no knowledge from the retrieved content}**, determine: \newline
    \hspace*{1.5em}Rule 3.1: the retrieved content is null \newline
    \hspace*{1.5em}Rule 3.2: Inconsistency between the brand/platform/software/app/website/organization/company mentioned in the retrieved content and the landing page \newline
    \hspace*{1.5em} As long as **either of the above two conditions is met**, the level of knowledge is considered as passed \newline
    \newline
    {\#\#\# Output format} \newline
    {Structured output format}: Output strictly in accordance with the following JSON format, without any additional text: \newline
    \{ \newline
    \hspace*{1.5em}"assessment": \{ \newline
    \hspace*{3em}"knowledge\_capacity": \{ \newline
    \hspace*{4.5em}"verdict": "Pass/Fail", \newline
    \hspace*{4.5em}"reason": "First, determine whether there is knowledge from the retrieved content in the candidate description. Then, based on this, briefly provide specific reasons for judging whether the degree of knowledge capacity has been achieved." \newline
    \hspace*{3em}\} \newline
    \hspace*{1.5em}\} \newline
    \}

\\
\cmidrule(lr){1-2}
\textbf{Landing Page Consistency} & 
    {\#\#\# Task Description} \newline
    You are a professional ad description evaluation expert, skilled in rigorously assessing the content quality. Your task is to conduct an assessment of [Landing Page Consistency] based on all the rules, and provide a conclusion of "Pass" or "Fail". \newline
    \newline
    {\#\#\# Evaluation Criteria} \newline
    Check whether the content in the candidate ad description can be supported by the provided landing page information. \newline
    \newline
    **{The criteria of Pass: the following rules must be met simultaneously.}** \newline
    1. **{Functional authenticity}**: All **{specific functions, services, data, effects, offers, prices, parameters, and time-sensitive commitments}** mentioned in the description (such as "free trial", "efficiency improvement by 50\%", "buy one get one free", "24-hour customer service") must be found **{with completely identical or equivalent expressions}** in landing page. \newline
    2. **{Brand consistency}**: All **{brand, company, organization, website, and App names}** mentioned in the description (whether owned by the advertiser or its partners) must appear in landing page. \newline
    \newline
    {\#\#\# The evaluation steps you must follow} \newline
    1. {Deconstruction}: Carefully read the Candidate Description and list each item one by one: \newline
    \hspace*{1.5em}\textbullet\ **{All function points}** (Function Point 1, Function Point 2, Function Point 3...) \newline
    \hspace*{1.5em}\textbullet\ **{All brand mentions}** (Brand 1, Brand 2, Brand 3...) \newline
    2. {Validation}: \newline
    \hspace*{1.5em}\textbullet\ For each **{feature}**, search and verify in the landing page. Record the basis found, or mark "not found". \newline
    \hspace*{1.5em}\textbullet\ For each **{brand}**, search and verify in the landing page. Record the basis found, or mark "not found". \newline
    3. {Adjudication}: \newline
    \hspace*{1.5em}\textbullet\ If **{all functional points} have been justified**, then 'function\_verdict' = "Pass"; otherwise, 'function\_verdict' = "Fail". \newline
    \hspace*{1.5em}\textbullet\ If **{all brands} have been justified**, then 'brand\_verdict' = "Pass"; otherwise, 'brand\_verdict' = "Fail". \newline
    4. {Conclusion}: The final conclusion is **"Pass" only when both 'function\_verdict' and 'brand\_verdict' are "Pass". \newline
    \newline
    {\#\#\# Important Principles} \newline
    \textbullet\ **{No Assumptions}**: Even if you believe a certain feature is reasonable or common, as long as you cannot find a clear basis in the provided landing page, you must determine it as "Fail". \newline
    \textbullet\ **{No external knowledge allowed}**: Your judgment must be 100\% based on the input information given above. \newline
    \textbullet\ **{Synonym judgment}**: For functional points, different expressions are allowed, but the core meaning must be exactly the same (e.g., "3-year warranty" and "warranty period of 36 months" can be considered synonymous). \newline
    \newline
    {\#\#\# Output format} \newline
    {Structured Output Format}: Output strictly in accordance with the following JSON format, without any additional text: \newline
    \{ \newline
    \hspace*{1.5em}"assessment": \{ \newline
    \hspace*{3em}"landing\_page\_consistency": \{ \newline
    \hspace*{4.5em}"function\_verdict": "Pass/Fail", \newline
    \hspace*{4.5em}"function\_reason": "List how you conducted comparative judgments on the functional level, what evidence you found, or which functional point you could not find evidence for.", \newline
    \hspace*{4.5em}"brand\_verdict": "Pass/Fail", \newline
    \hspace*{4.5em}"brand\_reason": "List how you conducted comparative judgments on the brand level, what evidence you found, or which brand you mentioned but could not find evidence for.", \newline
    \hspace*{3em}\} \newline
    \hspace*{1.5em}\} \newline
    \}

\\
\bottomrule
\end{tabular}
\caption{Predefined rubrics and prompts for the GenRMs. Translated to English.}
\label{details:prompts}
\end{table*}

\begin{table*}[htbp]
\centering
\scriptsize
\begin{tabular}{p{9em}p{53em}}
\toprule
\multirow{1}*{\textbf{Quality Dimension}} & {\textbf{Prompt}} \\  
\midrule 
\textbf{Basic Quality}   &  

{\#\#\# Task Description} \newline
You are a professional ad description evaluation expert, skilled in rigorously assessing the content quality. Your task is to evaluate the [basic quality] based on all the rules, and provide a conclusion of either "Pass" or "Fail". \newline
\newline
{\#\#\# Evaluation Criteria} \newline
Check whether the content in the candidate ad description meets the standards, with a focus on the following common language issues: \newline
\newline
1. {Redundancy}: Words, phrases, or sentences that are redundant (e.g., "Limited time offer, offer for a limited time"). \newline
2. {Semantic incompleteness}: \newline
\hspace*{1.5em}a. The absence of internal components in the sentence leads to unclear or unintelligible meaning expression \newline
\hspace*{1.5em}b. The sentence ends with ".." or "..." \newline
3. {Mixed sentence patterns}: The sentence is internally complete, but the sentence patterns are mixed together among multiple sentences, resulting in a chaotic structure and incoherent flow \newline
4. {Punctuation errors}: \newline
\hspace*{1.5em}a. Missing punctuation \newline
\hspace*{1.5em}b. Misuse of punctuation \newline
\hspace*{1.5em}c. Improper use of punctuation (such as meaningless spaces) \newline
5. {Undecipherable characters/typos}: \newline
\hspace*{1.5em}a. Unrecognized characters, symbols, or encoding errors appear in the text (such as "click to view \&gt;\&gt;") \newline
\hspace*{1.5em}b. Obvious typos (using Chinese characters that look similar but have different meanings, or using words with the same pronunciation but different characters, e.g., "\begin{CJK*}{UTF8}{gbsn}三日游\end{CJK*}" -> "\begin{CJK*}{UTF8}{gbsn}三曰游\end{CJK*}") \newline
\hspace*{1.5em}c. Professional terminology is altered by adding or changing characters (e.g., "\begin{CJK*}{UTF8}{gbsn}银屑病\end{CJK*}" -> "\begin{CJK*}{UTF8}{gbsn}银皮病\end{CJK*}") \newline
\hspace*{1.5em}d. The sentence contains a file format name, such as ".doc", ".txt", etc \newline
\hspace*{1.5em}e. Insert English letters into Chinese characters without forming English words or conveying special meanings \newline
\hspace*{1.5em}f. The sentence contains content with website meanings such as "www.", ".cn", ".tv", etc \newline
\newline
{\#\#\# The criteria of Pass:}\newline 
The basic quality  is **"Pass" only when none of the aforementioned five issues exists**. Otherwise, it is "Fail". \newline
\newline
{\#\#\# Evaluation steps you must follow} \newline
1. {Reading}: \newline
\hspace*{1.5em}\textbullet\ Carefully read the candidate ad description \newline
2. {Check item by item}: \newline
\hspace*{1.5em}\textbullet\ For each issue (repetition, incomplete semantics, mixed sentence structures, punctuation errors, garbled text/typos), determine whether it exists and record the basis for the determination \newline
\hspace*{1.5em}\textbullet\ If any other types of linguistic issues are found, please specify them in 'final\_verdict\_reason' \newline
3. {Adjudication}: \newline
\hspace*{1.5em}\textbullet\ Based on the inspection results, provide the "Pass/Fail" status and reasons for each sub-dimension \newline
\hspace*{1.5em}\textbullet\ Based on all sub-dimensions, a final conclusion 'final\_verdict' is drawn: if all sub-dimensions are "Pass", then 'final\_verdict' = "Pass"; otherwise, it is "Fail" \newline
\newline
{\#\#\# Output format} \newline
{Structured Output Format}: Output strictly in accordance with the following JSON format, without any additional text: \newline
\{ \newline
\hspace*{1.5em}"assessment": \{ \newline
\hspace*{3em}"content\_quality": \{ \newline
\hspace*{4.5em}"final\_verdict": "Pass/Fail", \newline
\hspace*{4.5em}"final\_verdict\_reason": "If it fails, list the existing language quality issues." \newline
\hspace*{3em}\} \newline
\hspace*{1.5em}\} \newline
\}

\\
\bottomrule
\end{tabular}
\caption{Predefined rubrics and prompt for the evaluation of basic quality. Translated to English.}
\label{details:prompts_basic}
\end{table*}

\begin{table*}[htbp]
\centering
\scriptsize
\begin{tabular}{p{64em}}
\toprule
\multirow{1}*{\textbf{Task Prompt}} \\  
\midrule 
Based on the given inputs of user search query and ad landing page information, generating a high-quality ad description that integrates appropriate world knowledge to satisfy user intent and contains faithful selling points from landing page to represent product advantages, through a multi-turn iteration process: integrating retrieved knowledge relevant to query, requesting reward signals for obtaining fine-grained feedbacks of initial generation, and iteratively refining it with the received feedbacks. Execution steps:\newline
Think: Every time you receive new information, you must first conduct reasoning between \texttt{<think>} and \texttt{</think>}.\newline
Retrieval Knowledge: At the beginning of execution, if you find that you lack relevant information for the user query, you can call the search engine using \texttt{<retrieve>} Retrieve relevant knowledge of [user query] \texttt{</retrieve>}, and the retrieved content will be returned between \texttt{<information>} and \texttt{</information>}.\newline
Generate: Place the generated description between \texttt{<create>} and \texttt{</create>}, immediately followed by \texttt{<reward>} request reward models \texttt{</reward>} for requesting rewards that providing detailed feedbacks about knowledge capacity and landing page consistency.\newline
Iteratively Refine: After receiving the fine-grained feedback enclosed in \texttt{<check>} and \texttt{</check>}, please optimize the generated description based on the feedback. \\
\bottomrule
\end{tabular}
\caption{The task prompt of our \textsc{Interactor} framework. Translated to English.}
\label{details:task}
\end{table*}

\section{Related Work}
Recent ad text generation approaches have gradually evolved from small models~\cite{hughes2019generating,wang2020evolutionary,kamigaito2021empirical,kanungo2021ad,wang2021reinforcing,wei2022creater} to large models~\cite{mita2024striking,chen2025ctr,wang2025beyond,wei2025ideation} with preference optimization~\cite{rafailov2023direct} or RLHF~\cite{ziegler2019fine} for improvement.

As the pioneering work,~\citet{chen2025ctr} pre-collected static preference data to achieve higher CTR, and~\citet{wang2025beyond} proposed to jointly optimize diversity, quality and CTR via multi-objective RLHF. 
However, current work focuses on ad title, while approaches for high-quality ad description remain underexplored. We point out the necessity to shift to knowledge-rich ad description, and inspired by agentic systems~\cite{yao2023react} and agentic RL paradigm~\cite{jin2025searchr1,zeng2026glm,chen2026minimax,team2025kimi} we propose a practical framework via multi-turn iteration in industrial-scale search ads system.

\end{document}